\documentclass{vldb}
\usepackage{graphicx}
\begin{document}

\title{\textbf{\large Clustering Co-occurrences of Maximal Frequent Patterns in Streams}}
\numberofauthors{3}
\author{
\alignauthor
Edgar H. de Graaf\\
       \affaddr{Leiden Institute of Advanced Computer Science\\ Leiden University}\\
       \affaddr{The Netherlands}\\
       \email{edegraaf@liacs.nl}
\alignauthor
Joost N. Kok\\
       \affaddr{Leiden Institute of Advanced Computer Science\\ Leiden University}\\
       \affaddr{The Netherlands}\\
       \email{joost@liacs.nl}
\alignauthor 
Walter A. Kosters\\
       \affaddr{Leiden Institute of Advanced Computer Science\\ Leiden University}\\
       \affaddr{The Netherlands}\\
       \email{kosters@liacs.nl}
}
\maketitle
\begin{abstract}
One way of getting a better view of data is by using frequent patterns. 
In this paper frequent patterns are (sub)sets that occur a minimal number of times in a 
stream of itemsets.
However, the discovery of frequent patterns in streams has always been problematic.
Because streams are potentially endless it is in principle impossible to say if a pattern
is often occurring or not. Furthermore, the number of patterns can be huge and
a good overview of the structure of the stream is lost quickly. The proposed approach 
will use clustering to facilitate the ``online'' analysis of the structure of the stream.

A clustering on the co-occurrence of patterns will give the user an improved view
on the structure of the stream. Some patterns might occur so often together that they should
form a combined pattern. 
In this way the patterns in the clustering will approximate the 
largest frequent patterns: maximal frequent patterns.
The number of (approximated) maximal frequent patterns is much smaller and combined with clustering methods 
these patterns provide a good view on the structure of the stream.

Our approach to decide if patterns occur often together is based on a method 
of clustering where only the distance between pairs of patterns is known. 
This distance is the Euclidean distance between points in a 2-dimensional space,
where the points represent the frequent patterns, or rather the most important ones.
The coordinates are adapted when the records from the stream pass by,
and reflect the real support of the corresponding pattern. In this
setup the support is viewed as the number of occurrences in a time window.
The main algorithm tries to maintain a dynamic model of the data stream
by merging and splitting these patterns.
Experiments show the versatility of the method.
\end{abstract}

\section{Introduction}


Effectively mining streams of data with \emph{frequent patterns}, i.e., patterns occurring at least a
minimal number of times, 
has always been a hard problem to tackle. The difficulty lies in the fact that you don't know which infrequent
patterns suddenly will become frequent and standard ways of pruning the search space are nearly impossible
to use. In this work \emph{patterns} are sets of items occurring in a record (also called transaction or \emph{itemset}) at a certain moment in time.

\medskip
\noindent \emph{Example 1}
Assume items $A$ and $B$ occur in every record, so they are frequent,
and therefore also the itemset $\{A, B\}$ will be frequent. However in stream context
we don't know they are frequent.
They may occur many times now and then never again. 

Furthermore, if $\{A, B, C\}$ is not frequent, $\{A, B, C, D\}$ and all other possible
additions to $\{A, B, C\}$ won't be frequent either. This well-known \emph{anti-monotone property}
is difficult to use in streams since $\{A, B, C\}$ may not be frequent only for a short
period. So $\{A, B, C\}$ in the stream as a whole might be frequent, but it doesn't seem
to be at the current moment since we have not seen all records. \hfill$\Box$

\medskip

One interesting application of frequent patterns is that they can be used to get an overview of the structure
of the dataset. Often there are too many patterns and further analysis of the patterns with for example clustering
is useful, especially in the case of streams where the set of frequent patterns is always changing. We will propose
a method of clustering where the distance between co-occurring maximal frequent subsets will be plotted in a 2-dimensional space.
\emph{Maximal frequent subsets} are sets of items occurring often in the stream while there is no frequently occurring
bigger set of items 
containing these same items. Each of these maximal frequent subsets represents a branch of subsets
occurring often together. If we combine this with information
about the distance between the maximal frequent subsets, then we can provide interesting structural information about the
stream. It will also possible to keep track of these sets
in an online way.


We will define our method of clustering and show its usefulness. To this end, this paper makes the
following contributions:
\\
--- We use a \textbf{dynamic support estimation} to determine the support
of those itemsets we need, and do this in an online way.
\\
--- It will be \textbf{explained how the distance between patterns is approximated},
using the supports, by pushing and pulling.
If this distance is large, patterns occur almost never together, and otherwise they 
do have many common occurrences.
\\
--- We will define when patterns \textbf{can be merged and when they should be split} to form smaller patterns 
and how this should be done. This could be considered as our major focus of interest.
\\
--- Finally through experiments \textbf{the effectiveness of our clustering is shown}
and efficiency is discussed.

We first mention related work, then we discuss the algorithm in full detail.
Finally we describe experiments and discuss these.

\section{Related Work}
This research is related to work done on clustering and in particular
clustering in streams. Also our work is related to frequent (maximal) pattern mining
in streams and large datasets. 

There are many algorithms for mining maximal frequent patterns, in ``normal'' datasets, in different ways.
We mention \textsc{GenMax} discussed in \cite{GoudaK} and \textsc{MAFIA} presented 
in \cite{BurdickD}.
Large datasets are different from streams in that there is an end to the dataset. One approach to mining
large datasets was proposed in \cite{El-HajjM}, where an extremely large
dataset is mined for maximal frequent patterns by proceeding in parallel.
Furthermore clustering on large datasets was
done in \cite{NanopoulosA}.
Much work has been performed on mining frequent patterns in (online) data streams, e.g., in \cite{ChangJH}
and \cite{JiangN}. In \cite{ChangJH2} frequent patterns are mined by using sliding window methods.
Our work has little overlap with work done on maximal pattern-based clustering as discussed 
in \cite{PeiJ} and \cite{WangH} where objects basically are clustered
by linking attribute groups with object groups when attributes have a minimal similarity. 
Related research has been done on clustering on streams in \cite{AggarwalCC}, where 
a study on clustering evolving data streams, (fast) changing data streams, is done.
Aggarwal et al. continue their work in \cite{AggarwalCC2} by clustering text and categorical
data in streams. Clustering categorical data was also done in \cite{GibsonD} where also
co-occurrence is used, but only for attribute values; the authors propose a visualization where the $x$-axis 
is the column position and the $y$-axis the distance based on co-occurrence of values.
Also in \cite{OCallaghanL} clustering on streams is mentioned, there the authors propose a new algorithm
and compare it with K-Means (see \cite{MacQueenJB}).

In this work a method of pushing and pulling points in accordance with a distance measure is used.
This technique was used before in \cite{CocxT} to cluster criminal careers and was 
developed in \cite{KostersWA}. This method of clustering was chosen since we only know the distance between
two patterns, where a low distance means frequent co-occurrence. We don't know the the precise $x$ and $y$ coordinates of 
the patterns, and therefore we cannot use standard methods of discovering clusters, e.g., K-Means.

\section{The Algorithm:\\Support and Distance,\\ Merge and Split}

Our goal is to produce an algorithm that is capable
of accepting a stream of records, each record being an unordered finite set of items,
meanwhile building a model of the maximal frequent itemsets.
The algorithm we propose, called \textsc{DistanceMergeSplit}, starts with randomly positioning $n$ points in a 2-dimensional space, e.g., in the unit square. Here $n$ is the
number of items maximally possible in an itemset. Each of these $n$ points represents one size 1 itemset, where the size of an itemset is
of course defined as the number of items it contains. These $n$ points remain present during the whole
process, though their coordinates may change. While the records from the data stream pass by,
new points are created (by merging or splitting) and 
others disappear (by merging, or by other reasons). Together these points constitute
the evolving model $\cal{P}$, where points correspond with frequent itemsets.

We will first explain how we use the stream
of records to update the supports of the elements of $\cal{P}$,
we then present an outline of the algorithm;
next we describe how the coordinates of the elements change in accordance
with the corresponding supports, and finally mention our method
of growing and shrinking the number of sets present in $\cal{P}$:
the merge and split part of the algorithm.

\subsection{Support}

The algorithm will receive a possibly infinite stream of itemsets, the records:
$r_1, r_2, r_3,\ldots$ Each time an itemset 
corresponding to a point in the space is a subset of a record, we observe an occurrence of this itemset.
We count the occurrences in the $m$ records we have seen so far
(and that can also be considered as the last $m$ records), and define support:
\begin{equation}\label{een} 
\mathit{support\;(p,m)} =  \sum_{t=1}^m \mathit{occurrence}\;(p,r_t) 
\end{equation}
\[\mathit{occurrence}\;(p, r) = \left\{ \begin{array}{l l} 1 & \quad \mbox{if $p \subseteq r$}\\
	0 & \quad \mbox{otherwise}\\
	 \end{array} \right. \]
Here $\mathit{p}$ is the pattern, the itemset, for which support is computed,
and $r$ is a record. 
If a new record arrives the support
needs to be adapted accordingly. 
Rather than using the full support for all records,
we will make use of a \emph{sliding window}
of size $\ell\geq 1$, and we will not keep the
data about the occurrences of the patterns in the transactions of this window.
Though this is not essential for 
our algorithm, it has a beneficial influence on the runtime, which
is especially interesting for an online algorithm.
If we have seen less than $\ell$ transactions ($m < \ell$) then we \emph{do}
use the previous formula
to calculate support.
This method will also be used when we later create new patterns online,
and is referred to as ``direct computation''; these patterns are then called ``young'',
as opposed to the ``old'' ones that are updated through equations~\ref{twee} and~\ref{drie} below. 
In the other case ($m \geq \ell$) we give an estimate $\mathit{support}_t(p)$
for the support during the last $\ell$ records in
the following way.
When the itemset $p$ is \emph{not} a subset of the current record 
$r_t$ we adapt the support as follows:
\begin{eqnarray}\label{twee}
\lefteqn{\mathit{support}_{t + 1}(p)} \ \ \ \ \ \ \ \ \\
&=&\!\!\mathit{support}_{t}(p)/\ell \cdot (\mathit{support}_{t}(p) - 1)\nonumber\\
&&+\ (1 - \mathit{support}_{t}(p)/\ell) \cdot \mathit{support}_{t}(p)\nonumber \\
&=&\!\! (1 - 1/\ell)\cdot\mathit{support}_{t}(p) \;\leq\; \mathit{support}_{t}(p)\nonumber
\end{eqnarray}
Indeed, when the first transaction of the window of size $\ell$ contains the pattern then support should decrease with 1. However, 
if the first record also does not contain $p$, then support remains the same. It is important to notice that the probability 
of a transaction containing $p$ in a window of size $\ell$ is estimated with $\mathit{support}_{t}(p)/\ell$.
If the new record \emph{does} contain the itemset $p$ then support is adapted as follows:
\begin{eqnarray}\label{drie}
\lefteqn{\mathit{support}_{t + 1}(p)} \ \ \ \ \ \ \ \ \\
&=&\!\!\mathit{support}_{t}(p)/\ell \cdot \mathit{support}_{t}(p)\nonumber\\ 
& &+\ (1 - \mathit{support}_{t}(p)/\ell) \cdot (\mathit{support}_{t}(p) + 1)\nonumber\\
&=&\!\! (1 - 1/\ell) \cdot\mathit{support}_{t}(p)+ 1\; \geq\;\mathit{support}_{t}(p)\nonumber
\end{eqnarray}
Now when the first transaction of the window of size $\ell$ contains the pattern then support remains unchanged as
the window shifts. However, if it does not contain the pattern $p$, then support will increase with 1.
Both formulas \emph{assume that occurrences are uniformly spread} over the window of size $\ell$, but by 
using these formulas to adapt support we do not have to keep all occurrences for all patterns in the 2-dimensional space.
Notice that $0\leq\mathit{support}_{t}(p)\leq\ell$ always holds.

We have now described how the stream of records influences the supports of the itemsets
that are currently being tracked, i.e., those in $\cal{P}$. Note that the itemsets of size 1 are 
always present in $\cal{P}$, for reasons mentioned in the next paragraph.
Larger itemsets may appear and disappear as the algorithm proceeds.
Also observe that the supports are estimates, due to the application
of equations~\ref{twee} and~\ref{drie}.

\subsection{The Algorithm}
The algorithm works with the set $\cal{P}$ of patterns that are currently
present, represented by (coordinates of) points in 2-dimensional
Euclidean space.
The outline of the algorithm \textsc{DistanceMergeSplit} is as follows:
\vspace*{5mm}\hrule\vspace*{-1mm}
\begin{tabbing}
XX\=XX\=XX\=XX\=XX\=XX\kill
\>initialize $\cal{P}$ with the $n$ itemsets of size 1\\
\>\textbf{for} $t\leftarrow 1$ to $\infty$ \textbf{do}\\
\>\>$\cal{Q}\leftarrow\emptyset$\\
\>\>\textbf{for} all patterns $p\in\cal{P}$ \textbf{do}\\
\>\>\>compute $\mathit{support}_t(p)$ using the $t^{\mathrm{th}}$ record $r_t$,\\
\>\>\>either through updating (old patterns)\\
\>\>\>or by direct computation (young ones)\\
\>\>\textbf{for} a random subset of pairs of patterns in $\cal{P}$ \textbf{do}\\
\>\>\>update their distance according to their support\\
\>\>\textbf{for} all ``appropriate'' pattern pairs in $\cal{P}$ \textbf{do}\\ 
\>\>\>\emph{merge} the pair, creating (new) pattern(s) in $\cal{Q}$\\
\>\>\>mark the smallest of the pair,\\
\>\>\>or both if their sizes are equal\\
\>\>remove the marked patterns from $\cal{P}$\\
\>\>\textbf{for} all patterns $p\in\cal{P}$ \textbf{do}\\
\>\>\>\textbf{if} $p$ is infrequent and old enough \textbf{then}\\
\>\>\>\>\emph{split} $p$ into (new) patterns in $\cal{Q}$\\
\>\>\>\>remove $p$ from $\cal{P}$\\
\>\>${\cal{P}} \leftarrow \cal{P}\cup\cal{Q}$, joining duplicates\\
\>\>remove non-maximal frequent patterns from $\cal{P}$
\end{tabbing}
\vspace*{-1mm}\hrule\vspace*{1mm}
\centerline{\textsc{DistanceMergeSplit}}
\vspace*{1.2mm}
\hrule\vspace*{5mm}

Note that itemsets of size 1 are \emph{never} removed from $\cal{P}$,
not even when they are infrequent.
The size 1 itemsets are always present, and play a special role: besides the fact that some of them are frequent, they also serve as building blocks. In many cases they are not maximal. If they were removed,
it could be impossible to re-introduce single items after having become infrequent.

Patterns that are new in $\cal{P}$ are called ``young''.
When computing supports for these patterns, we use equation~\ref{een},
when updating the ``old'' ones we use equations \ref{twee} and \ref{drie}.
So, each pattern present in $\cal{P}$ also has an \emph{age}:
patterns that have an age smaller than the window size $\ell$
are ``young'', the others are ``old''.

On two occasions the algorithm introduces indeterminism:
first, when the support computation is done using the approximating updates
for ``old'' patterns (saving a lot of time and memory)
and second, when pushing and pulling a random subset of the pairs, see below.

\subsection{Distance}
We now describe how the coordinates of the points change as
their supports vary when the new records from the stream come in.
In our model for $\mathit{distance}\;(p_{1},p_{2})$ we take the Euclidean distance between the 2-dimensional coordinates of the points corresponding with the
two patterns $p_1$ and $p_2$. These points are pulled closer to one another
if they occur in the current transaction and they
are pushed apart if not. Furthermore nothing is done if both do not occur.
In every time step a random selection of the pairs undergoes this process.

To pull two points together we set the \emph{goal distance} to 0 and to push them apart the goal distance is $\sqrt{2}$, which is the maximum Euclidean distance between any two
points in the unit square.
These distances are then used to update the coordinates 
$(x_{p_{1}},y_{p_{1}})$ and $(x_{p_{2}},y_{p_{2}})$  of the points:
\begin{enumerate}
	\item $x_{p_{1}} \leftarrow x_{p_{1}} - \alpha \cdot (\mathit{distance}\;(p_{1},p_{2}) - \gamma) \cdot (x_{p_{1}} - x_{p_{2}})$
	\item $y_{p_{1}} \leftarrow y_{p_{1}} - \alpha \cdot (\mathit{distance}\;(p_{1},p_{2}) - \gamma) \cdot (y_{p_{1}} - y_{p_{2}})$
	\item $x_{p_{2}} \leftarrow x_{p_{2}} + \alpha \cdot (\mathit{distance}\;(p_{1},p_{2}) - \gamma) \cdot (x_{p_{1}} - x_{p_{2}})$
	\item $y_{p_{2}} \leftarrow y_{p_{2}} + \alpha \cdot (\mathit{distance}\;(p_{1},p_{2}) - \gamma) \cdot (y_{p_{1}} - y_{p_{2}})$
\end{enumerate}
Here $\alpha$ ($0 \leq \alpha \leq 1$) is the user-defined learning rate and $\gamma$ ($0 \leq \gamma \leq \sqrt{2}$) is the goal distance. 

These formulas are basically the same as the one defined in \cite{KostersWA}, however we use the distances
to \emph{decide when to merge}.
Points may leave the unit square; however, when presenting the results of
the experiments, such points are projected on the nearest wall
of this square.

\subsection{Merge and Split}

Now we describe how we merge and split the itemsets of the model as time goes by. The cluster model $\cal{P}$ contains points with
corresponding itemsets. When the distance between two points is small, then the corresponding itemsets occur many times together.
In some cases one itemset can be made that represents two of them: the algorithm will try these combinations.
For some combinations it is possible that they turn out to be not so good, their frequency is smaller than $\mathit{minsupp}$, where $\mathit{minsupp}$ is a user-defined threshold.
This can happen when their combined frequency is lower than $\mathit{minsupp}$ or suddenly frequency drops below $\mathit{minsupp}$.
In either case we need to split the size $k$ itemset into $k$ itemsets of size $k - 1$,
all being subsets of the original itemset.
Later we will discuss splitting in more detail, we now first explain merging. 

As transactions come in, some of the initial size 1 itemsets become \emph{frequent}, meaning that the support is higher than $\mathit{minsupp}$. These sets can --- under certain circumstances, see below ---
merge to itemsets of size 2, and so on: two itemsets $p_1$ and $p_2$ are merged if (in the algorithm above the following series of conditions is referred to
as ``appropriate''):

\begin{itemize}
  \item The two itemsets $p_1$ and $p_2$ currently are frequent, i.e., 
  it holds that both $\mathit{support}_t(p_{1}) \geq \mathit{minsupp}$ and $\mathit{support}_t(p_{2}) \geq \mathit{minsupp}$. (Note that this condition automatically holds for all (pairs of) itemsets in $\cal{P}$ that have size larger than 1.)
  \item The itemsets are close together in the model, so they (probably) occur often together as a subset of transactions in the stream: $\mathit{distance}\;(p_{1},p_{2}) \leq \mathit{mergedist}$,
  where $\mathit{mergedist}$ is a user-defined threshold for the distance between $p_{1}$ and $p_{2}$ below which merging is allowed.
  \item 
  The pattern $p_2$ has an item $i_{p}$ which is not in the pattern $p_1$, such that $p_2\setminus\{i_{p}\}\subseteq p_1$. (This condition always holds if $p_2$ has size 1.)
  \item
  The patterns $p_1$ and $p_2$ are old enough: they exist in $\cal{P}$ for at least
  $\ell$ (the window size) records. (Note that the supports of these sets are currently updated through
  equations~\ref{twee} and~\ref{drie} above.)
\end{itemize}

If the patterns $p_1$ and $p_2$ are of equal size then for merging we create the set 
$p_1\cup p_2$.
Both original patterns are removed from the 2-dimensional space except if their size is 1.

\medskip
\noindent\emph{Example 2}
Say $p_{1} = \{A, B, C\}$ and $p_{2} = \{A, B, D\}$. Furthermore, suppose 
$\mathit{distance}\;(p_{1},p_{2}) = 0.1$ and $\mathit{mergedist} = 0.2$. 
Then the new itemset $q= \{A, B, C, D\}$ is added to the cluster model $\cal{Q}$
(and later to $\cal{P}$)
with a randomly chosen $x$ and $y$ position. Both $p_{1}$ and $p_{2}$ are removed from $\cal{P}$ after all merging is done. 
It could be the case that $\{B, C, D\}$ and/or $\{A, C, D\}$ is infrequent, implying that 
$\{A, B, C, D\}$ will be infrequent too. However, in that case $\{A, B, C, D\}$
will disappear due to splitting; the patterns $p_1$ and $p_2$ should not have been
so close together in the first place.
\hfill$\Box$

\medskip

If pattern $p_{1}$ contains more items than $p_{2}$ and $p_{2}\setminus\{i_p\}\subseteq p_1$
for some $i_p\in p_2$ with  $i_p\not\in p_1$, then
for each item $e \in p_{1} \setminus p_{2}$ we add an itemset $p_{2} \cup \{e\}$. This enables patterns to be merged
with patterns that already were merged before and disappeared from the model. The smaller pattern $p_{2}$ 
is removed except if it is of size 1.

\medskip
\noindent\emph{Example 3}
Assume
$p_{1} = \{A, B, C, D, E\}$, $p_{2} = \{A, B, F\}$, $\mathit{distance}\;(p_{1},p_{2}) = 0.1$ and
$\mathit{mergedist} = 0.2$. The algorithm will add $\{A, B, F, C\}$,
$\{A, B, F, D\}$ and $\{A, B, F, E\}$ to
$\cal{Q}$ (and later to $\cal{P}$). All $x$ and $y$ positions of the corresponding points are
again randomly chosen. The itemset $p_{2}$ is removed from $\cal{P}$ after all merging is done, $p_1$ stays in $\cal{P}$.
\hfill$\Box$

\medskip






%

Next we \emph{split} patterns, when they contain more than one item, if they do not occur often enough
and they have been in the model for at least a certain number of records
(they are ``old enough''). Split combinations are generated
by removing each item from the original pattern once. 
The remaining items form one new itemset, so in this way a size $k$
itemset will result in $k$ combinations after splitting.

\medskip
\noindent\emph{Example 4}
Assume
$p = \{A, B, F\}$ has support $<\mathit{minsupp}$,
and exists long enough in $\cal{P}$. The algorithm will add $\{A, B\}$,
$\{A, F\}$ and $\{B, F\}$ to
$\cal{Q}$ (and later to $\cal{P}$),
located at random points.
The itemset $p$ is removed from $\cal{P}$.
\hfill$\Box$

\medskip

Finally, the newly formed patterns in $\cal{Q}$ are 
united with those in $\cal{P}$. Of course,
when patterns occur more than one time, only one copy --- the oldest one ---
is maintained. And those patterns from $\cal{P}$ that are contained in a larger
one in $\cal{P}$ are removed, unless --- as stated above --- they have size 1:
we focus on the maximal patterns.


\section{Experiments and Discussion}

The experiments are organized such that we first show the method at work in a few controlled synthetic cases.
Then we will use the algorithm to build a cluster model for a real dataset, showing some ``real life'' results.
The first synthetic experiment will be a stream with 10 groups of 5 items. Groups do not occur together, but all of them
occur often. This dataset is called the \emph{10-groups dataset}. The second synthetic experiment will be a stream where certain groups of 
items suddenly do not occur; instead another group starts occurring.
We call this dataset the \emph{sudden change dataset}. 
Finally one experiment will take the stream of the first experiment and
it will test the effect of different noise levels; it will be called the \emph{noise dataset}.
The \emph{real dataset} is the Large Soybean Database used for soybean disease diagnosis in \cite{MichalskiRS}.
The dataset contains 683 records with 35 attributes. First we removed all missing values and we converted each record
to a string of $n=84$ yes/no values for each attribute value. In this research we do not deal
with missing values, and each
item represents an attribute value.

All experiments were performed on an Intel Pentium 4 64-bits 3.2 Ghz machine with 3 GB memory. As operating system Debian
Linux 64-bits was used with kernel 2.6.8-12-em64t-p4.

\begin{figure}[!ht]
\includegraphics[width=85mm]{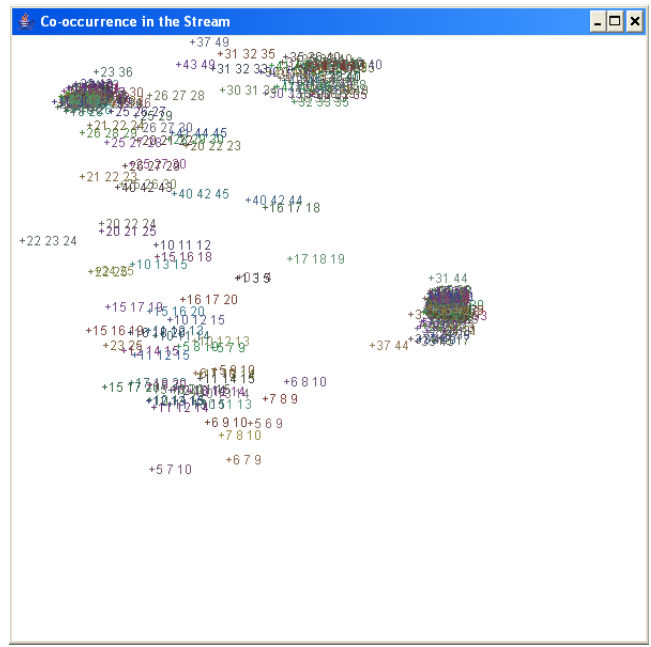}
\caption{Model after seeing 600 transactions of the 10-groups dataset
($n = 50$, $\mathit{minsupp} = 15$, $\ell=\mathit{window\ size} = 300$, $\mathit{mergedist} = 0.1$, $\alpha = 0.1$).}\label{fig:exp1a}
\end{figure}

\begin{figure}[!ht]
\includegraphics[width=85mm]{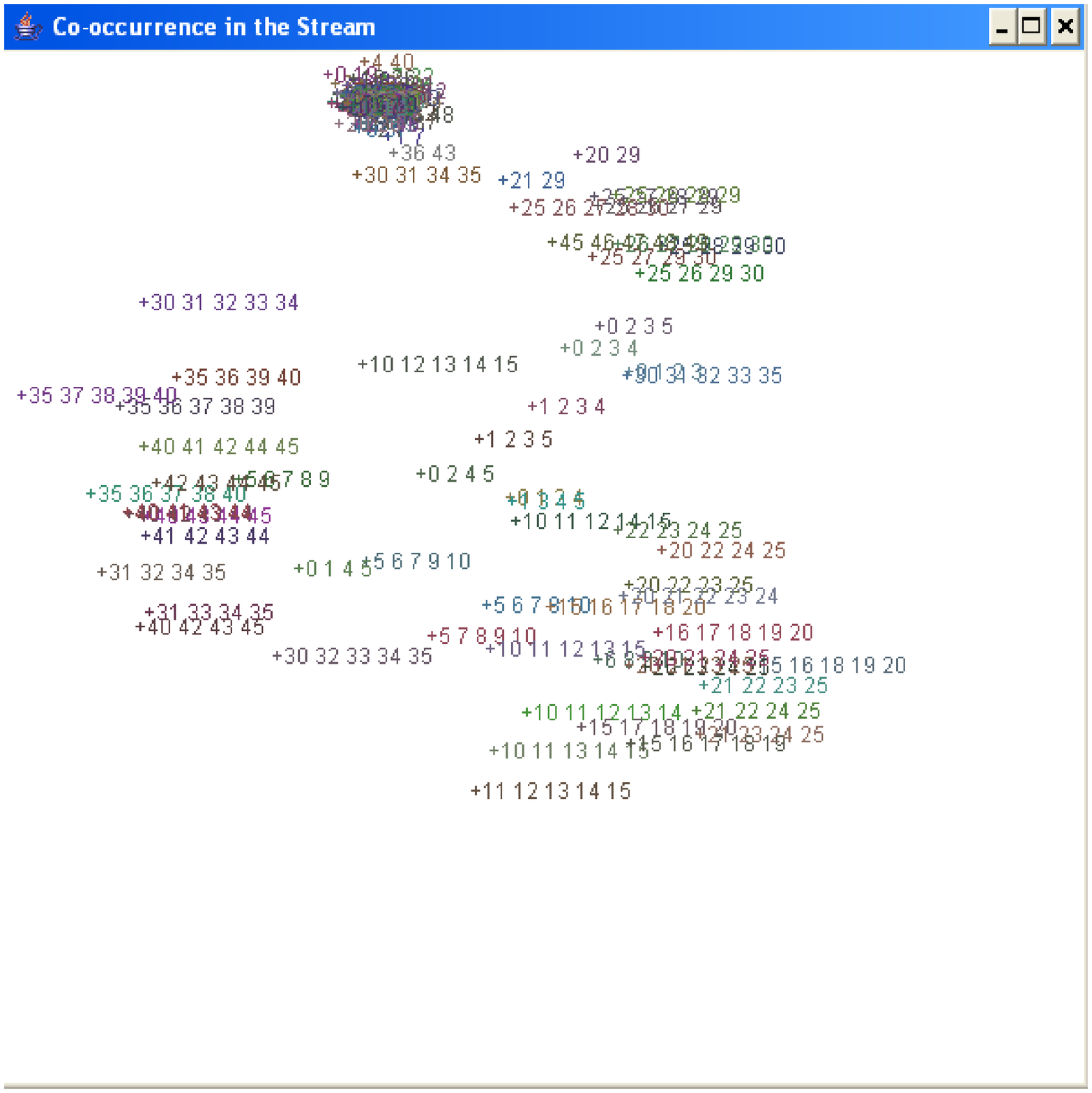}
\caption{Model after seeing 1,200 transactions of the 10-groups dataset
($n = 50$, $\mathit{minsupp} = 15$, $\ell=\mathit{window\ size} = 300$, $\mathit{mergedist} = 0.1$, $\alpha = 0.1$).}\label{fig:exp1b}
\end{figure}

\begin{figure}[!ht]
\includegraphics[width=85mm]{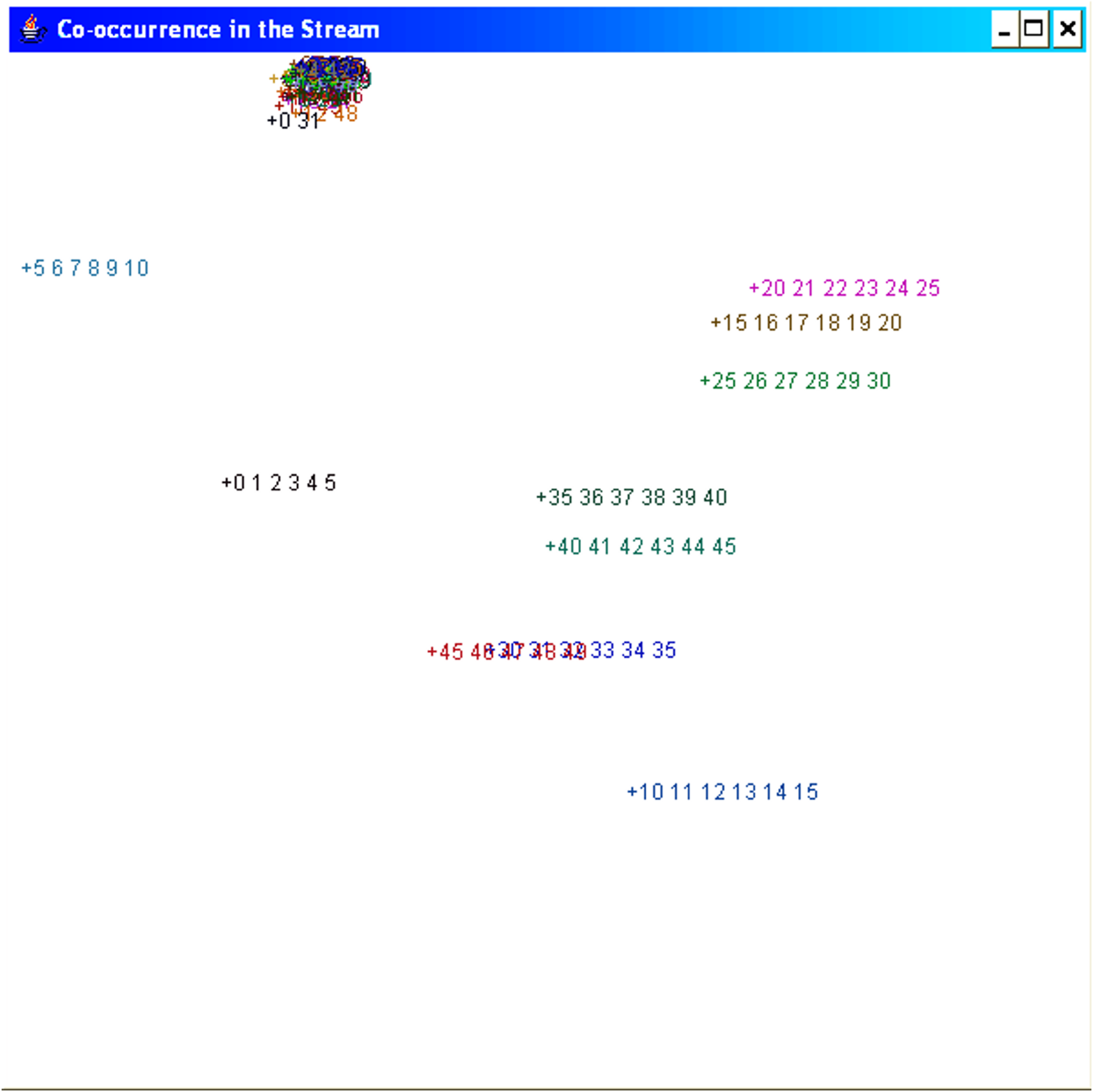}
\caption{Model after seeing 4,500 transactions of the 10-groups dataset
($n = 50$, $\mathit{minsupp} = 15$, $\ell=\mathit{window\ size} = 300$, $\mathit{mergedist} = 0.1$, $\alpha = 0.1$).}\label{fig:exp1c}
\end{figure}

\begin{figure}[!ht]
\includegraphics[width=85mm]{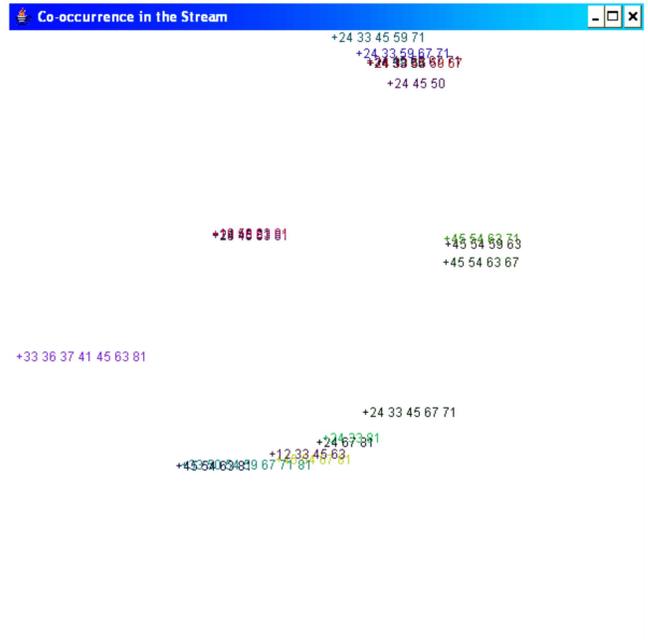}
\caption{Model after 20,000 transactions of the real dataset were processed 
($n = 84$, $\mathit{minsupp} = 120$, $\ell=\mathit{window\ size} = 300$, $\mathit{mergedist} = 0.1$, $\alpha = 0.1$).}\label{fig:exp4}
\end{figure}

Figures \ref{fig:exp1a}, \ref{fig:exp1b} and \ref{fig:exp1c} show how the cluster model changes as more transactions
are coming in for the 10-groups dataset. The first group of this dataset consists of items 0 to 5, the second has 5 to 10, etc. In the last
figure, Figure~\ref{fig:exp1c}, we clearly see these patterns. Furthermore notice that both the second and the first group contain
the item 5, so there is a slight overlap. We see these itemsets closer together because they are both close to the pattern $\{5\}$.
In order to get a clear picture we did not display the size 1 itemsets.
Itemsets are plotted using $+$s, accompanied by the items they contain.

Figure~\ref{fig:exp4} displays the cluster model (only patterns with age at least 50 are shown) 
after seeing 20,000 transactions produced by repeating the real dataset.
Some patterns, i.e., itemsets, are clearly placed far apart from each other or close together. 
Table~\ref{table:exp42} displays some examples
on the co-occurrences of patterns. The first thing to notice is that all the patterns occur often and so they should be in
the cluster model. Secondly the first and the second itemset 
occur often together, so we expect them to be close together in
the model. Finally the last itemset does not occur less often with the other two, we expect them to be placed further apart. Figure~\ref{fig:exp4} 
displays all these facts in one picture.

\begin{table}[!ht]
 \begin{center}
 \begin{tabular}{|l|c|c|c|}
  \hline
							& \{24, 33, 81\}	& \{24, 67, 81\} & \{24, 45, 50\} \\
 \hline
\{24, 33, 81\} &			295/683				& 253/683				 & 182/683 \\
 \hline
\{24, 67, 81\} &			253/683				& 260/683				 & 189/683 \\
 \hline
\{24, 45, 50\} &			182/683				& 189/683				 & 237/683 \\
 \hline
 \end{tabular}
\end{center}
 \caption{Three patterns from the model of Figure \ref{fig:exp4} and their co-occurrence.}\label{table:exp42}
 \end{table}

\begin{figure}[!ht]
\includegraphics[width=85mm]{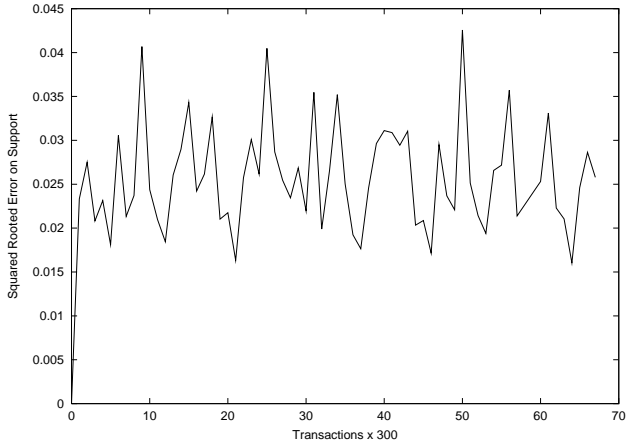}
\caption{The root squared error between the real support and the approximated support in the time window for the real dataset
($n = 84$, $\mathit{minsupp} = 150$, $\ell=\mathit{window\ size} = 300$, $\mathit{mergedist} = 0.1$, $\alpha = 0.1$).}\label{fig:error}
\end{figure}

Approximating supports well is important in order to know which itemsets should be split. In Figure~\ref{fig:error} we 
show for all patterns in a computed cluster model, with a minimal age of 300, the error between their approximated support and their real support 
in the time window as the transactions from the real dataset arrive. The root mean squared error of the supports for this
model is never larger than $0.045$. All supports are first made relative to the time window size by dividing by $300$.

Our cluster model is said to approach the maximal frequent patterns. In order to show that it is able to do so, we first extracted
from the original real dataset (683 transactions) all maximal frequent 
patterns using the \textsc{Apriori} algorithm 
with $\mathit{minsupp} = 341$, which corresponds to a relative support of $0.5$. Then we produced a model 
where $\ell=\mathit{window\ size} = 1,000$, $\mathit{minsupp} = 500$, $\mathit{mergedist} = 0.1$ and $\alpha = 0.1$.
In Table~\ref{table:exp41} some statistics are shown.
\begin{table}[!ht]
\begin{center}
\begin{tabular}{|l|c|}
 \hline
 Number of exactly matching patterns & 19 out of 45 \\
  \hline
 Number of patterns with zero or one & \\
 items extra & 35 out of 45\\
 \hline
 Number of patterns not in the model & 10 out of 45\\
 \hline
 Root squared error & \\
 for the relative support of & 0.0176\\
 matching maximal frequent patterns & \\
 \hline
 \end{tabular}
\end{center}
 \caption{The approximation of the maximal frequent patterns in 
 the real dataset after seeing 20,000 transactions
 ($n = 84$, $\ell=\mathit{window\ size} = 1,000$, $\mathit{minsupp} = 500$, $\mathit{mergedist} = 0.1$, $\alpha = 0.1$).}\label{table:exp41}
 \end{table}

Many of the maximal frequent patterns exist in the model, however the algorithm constantly tries 
extending itemsets based on an approximated distance. Because of this the model contains the
maximal frequent patterns with an extra item. As a future improvement we might keep all
itemsets until their superset is not young any longer.
Only a few itemsets do not exist in the model, but many of their subsets were found.
The root squared error for the 19 matching patterns is about $0.0176$.

 \begin{figure}[!ht]
\includegraphics[width=85mm]{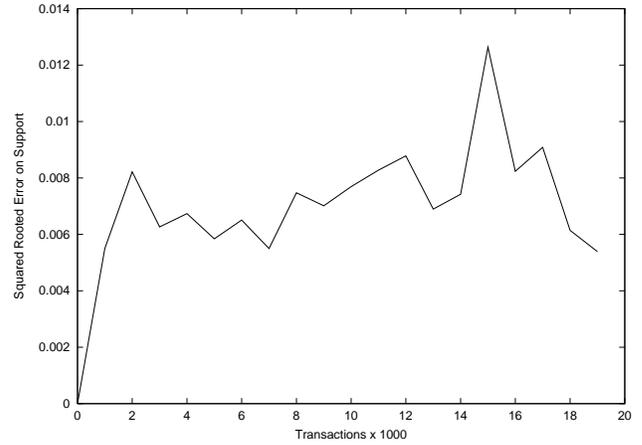}
\caption{The root squared error between the real support and the approximated support in the time window for the real dataset
($n = 84$, $\mathit{minsupp} = 500$, $\ell=\mathit{window\ size} = 1,000$, $\mathit{mergedist} = 0.1$, $\alpha = 0.1$).}\label{fig:error2}
\end{figure}

The bigger time window used in the experiment of Figure~\ref{fig:error2} shows a small improvement for the root squared error.

\begin{figure}[!ht]
\includegraphics[width=85mm]{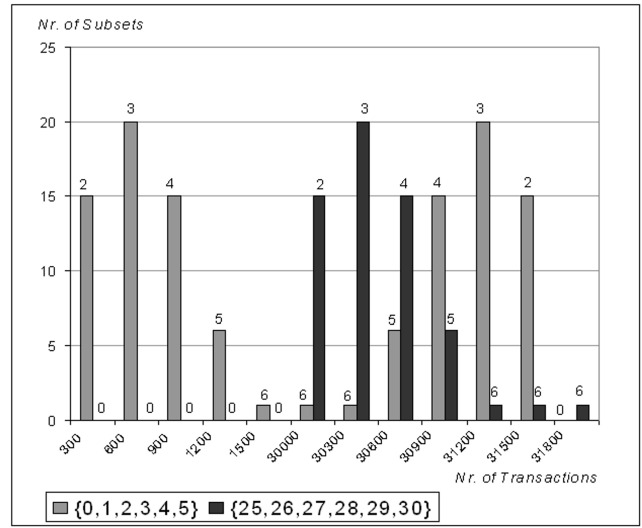}
 \caption{The sudden change dataset, the stream changes in the middle
 ($n = 50$, $\mathit{minsupp} = 15$, $\ell=\mathit{window\ size} = 300$, $\mathit{mergedist} = 0.1$, $\alpha = 0.1$).}\label{fig:exp2}
\end{figure}

 The second synthetic dataset, called the sudden change dataset, simulates a stream that completely changes
 after seeing many transactions (i.e., 30,000). The results are displayed in Figure~\ref{fig:exp2}, where the
 labels above each bar reveal the size of the itemsets. First the records in the stream
 always contain items 1 to 5. Then after 30,000 transactions they only contain items 25 to 30. Figure~\ref{fig:exp2}
 shows how the first pattern appears and how it slowly disappears in the middle and in the end the model contains
 only the patterns with items 25 to 30.

Finally Table~\ref{table:exp3} shows how noise influences the results. In the noise
dataset each time a group of the same 11 items appears, first items 0 to 10, 10 to 20,
etc. If the noise level is $r$\;\%, then approximately $r$\;\% of the items will not appear
even though they should have. Table~\ref{table:exp3} shows that, even if there is noise, 
the correct itemsets are generated at least in part after seeing 50,000 transactions. We call 
an itemset \emph{correct} if we would \emph{expect} it. If a group contains items 0 to 10 then we would
expect to see subsets with items 0 to 10. However \emph{unexpected} would be to see itemsets
with items 0 to 10 and some items outside this range. These unexpected subsets (subsets of all
items in the group) did not occur often and their size was never bigger than 4 items.

\begin{table}
\begin{center}
\begin{tabular}{|c|c|}
\hline
Noise probability & Number of\\
of items (\%) & expected subsets\\
 \hline
0		& 5\\
 \hline
10		& 3\\ 
 \hline
20		& 28\\
 \hline
30		& 19\\
 \hline
40		& 10\\
\hline
\end{tabular}

\vspace*{5mm}

\begin{tabular}{|c|c|}
\hline
Size range of & Number of\\
expected subsets &unexpected subsets\\
\hline
10 to 11 (items) & 0\\
\hline
10 to 11 & 2\\ 
\hline
5 to 6 & 0\\
\hline
3 to 4 & 1\\
\hline
2 to 3 & 4 \\
\hline
\end{tabular}

\end{center}
 \caption{The noise dataset, where the influence of noise on the structure is shown
 ($n = 50$, $\mathit{minsupp} = 15$, $\ell=\mathit{window\ size} = 300$, $\mathit{mergedist} = 0.1$, $\alpha = 0.1$).}\label{table:exp3}
\end{table}

\begin{figure}[!ht]
\hfill
\begin{minipage}[t]{.49\textwidth}
\begin{center}
\includegraphics[width=6cm]{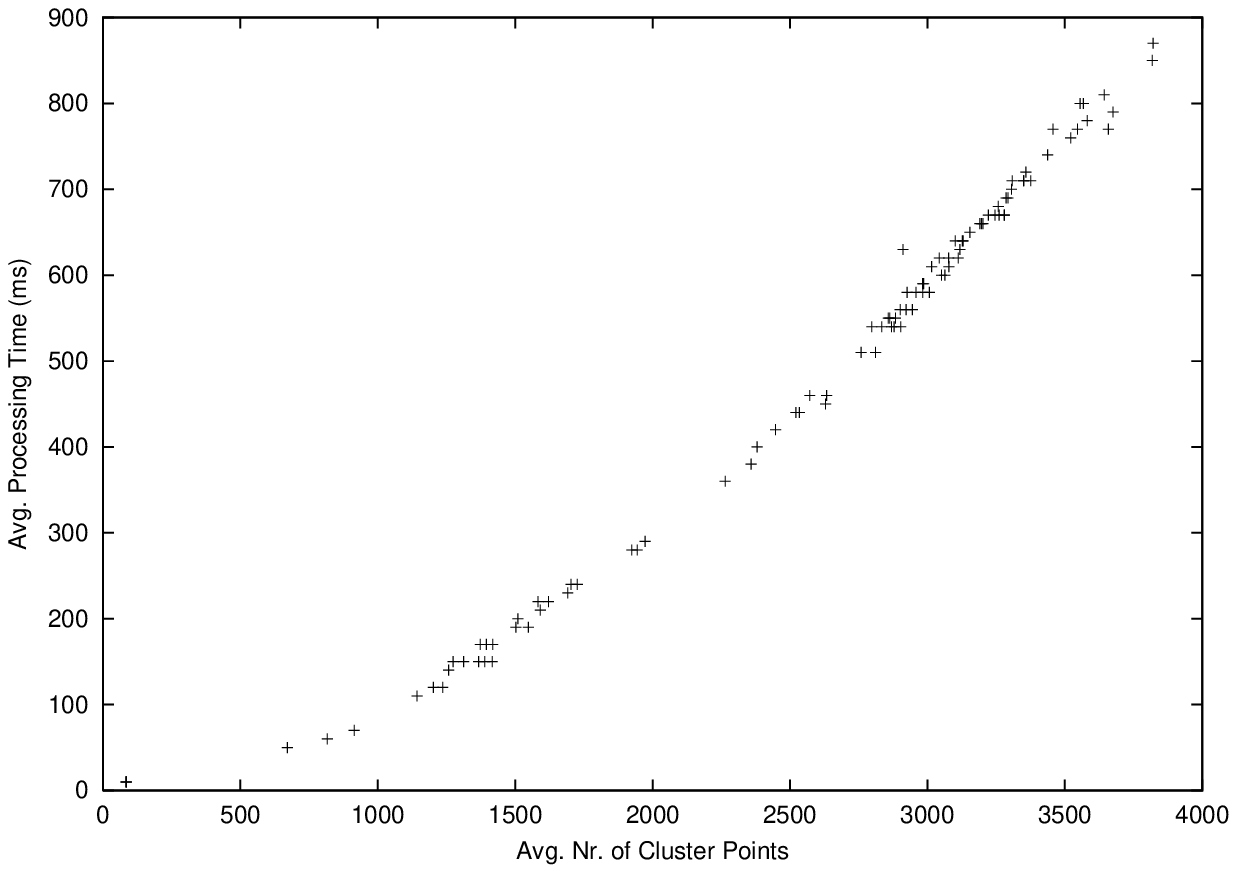}
\caption{Transaction processing time in milliseconds for different cluster model sizes for the real dataset
($n = 84$, $\mathit{minsupp} = 60$, $\ell=\mathit{window\ size} = 300$, $\mathit{mergedist} = 0.1$, $\alpha = 0.1$).}\label{fig:pointsms}
\end{center}
\end{minipage}
\hfill

\vspace*{5mm}
\begin{minipage}[t]{.49\textwidth}
\begin{center}
\includegraphics[width=6cm]{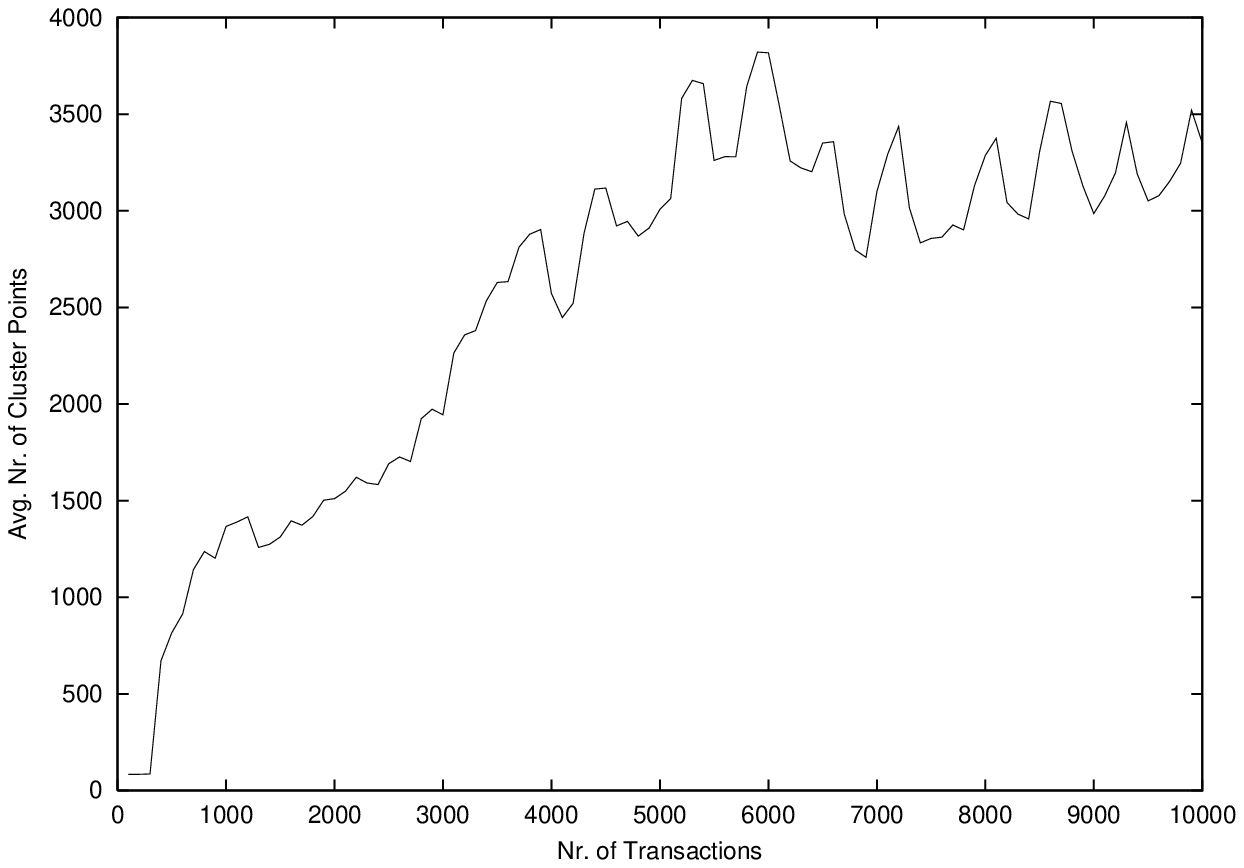}
\caption{Development of cluster model size as transactions of the real dataset are processed
($n = 84$, $\mathit{minsupp} = 60$, $\ell=\mathit{window\ size} = 300$, $\mathit{mergedist} = 0.1$, $\alpha = 0.1$).}\label{fig:pointstrans}
\end{center}
\end{minipage}
\hfill
\end{figure}

The processing time of the algorithm strongly depends on the support threshold $\mathit{minsupp}$ one chooses. The lower $\mathit{minsupp}$ is chosen
the more points the cluster model will contain eventually and so processing time will get longer. 
Figure~\ref{fig:pointsms} shows that the average processing time for each transaction gets worse as the model contains
more itemset points. However, Figure~\ref{fig:pointstrans} shows that, for the real dataset, the number of 
points in the model eventually stabilizes. For each transaction we adapt the distances between points a number
of times. In the case of the real dataset we randomly choose pairs 40,000 times in order to push or pull them, depending
on their co-occurrence. Obviously one way of speeding up processing is to make it less than 40,000 times or one
can skip adapting distances sometimes.

\section{Conclusions and Future Work}


The algorithm presented in this paper will generate a cluster model of the maximal frequent itemsets and their co-occurrences.
This gives the user a quick view on the patterns, frequent subsets, in the stream and how they occur in the stream. E.g., a
shop keeper will know which products are often sold together and for the groups of products not often sold together
the model indicates how much they are not sold together.

The co-occurrence distance of patterns is computed by pushing apart or pulling together patterns in a 2-dimensional space. 
Pushing was done when only one of the patterns occurs and pulling if they occur together. This distance is used to merge 
patterns together if it is smaller than a user-defined threshold, because we want only maximal frequent itemsets (itemsets that
are often a subset of a transaction but they are never a subset of a bigger frequent itemsets) such that the model does not grow too big.
Finally points are split if they happen to occur less than expected. Splitting and merging is required because the cluster model cannot
contain all pattern since in streams we never know which items are frequent due to its possible infinite nature.

In the future we want to focus more on the applications of our algorithm and how it is best used in the analysis of streams. Furthermore we would like to examine how well
the support estimates are, and how extra parameters (e.g., to determine the threshold
age for splitting)
can be employed.

\section{Acknowledgment}
This research is carried out within the Netherlands Organization for Scientific Research (NWO) MISTA Project (grant no. 612.066.304).

\bibliographystyle{abbrv}
\bibliography{maart19}

\end{document}